\documentclass{article}

     \PassOptionsToPackage{numbers, compress}{natbib}
     \usepackage[final]{neurips_2019}

\usepackage[utf8]{inputenc} % allow utf-8 input
\usepackage[T1]{fontenc}    % use 8-bit T1 fonts
\usepackage{hyperref}       % hyperlinks
\usepackage{url}            % simple URL typesetting
\usepackage{booktabs}       % professional-quality tables
\usepackage{amsfonts}       % blackboard math symbols
\usepackage{nicefrac}       % compact symbols for 1/2, etc.
\usepackage{microtype}      % microtypography
\usepackage{todonotes}

\usepackage{amsmath}
\usepackage{mathtools}
\usepackage[linesnumbered,ruled]{algorithm2e}
\usepackage{multirow}
\usepackage{makecell}

\usepackage{xcolor}
\usepackage[normalem]{ulem}
\newcommand\redsout{\bgroup\markoverwith{\textcolor{red}{\rule[0.5ex]{2pt}{0.4pt}}}\ULon}

\newcommand{\vpara}[1]{\vspace{0.1in}\noindent\textbf{#1 }}

\newcommand{\reals}{\mathbb{R}}

\title{Graph Representation Learning via Multi-task Knowledge Distillation}

\author{%
    Jiaqi Ma\\
    School of Information \\
    University of Michigan \\
    \texttt{jiaqima@umich.edu} \\
    \And
    Qiaozhu Mei \\
    School of Information\\
    University of Michigan \\
    \texttt{qmei@umich.edu} \\
}

\begin{document}

\maketitle

\begin{abstract}
Machine learning on graph structured data has attracted much research interest due to its ubiquity in real world data. However, how to efficiently represent graph data in a general way is still an open problem. Traditional methods use handcraft graph features in a tabular form but suffer from the defects of domain expertise requirement and information loss. Graph representation learning overcomes these defects by automatically learning the continuous representations from graph structures, but they require abundant training labels, which are often hard to fulfill for graph-level prediction problems.  In this work, we demonstrate that, if available, the domain expertise used for designing handcraft graph features can improve the graph-level representation learning when training labels are scarce. Specifically, we proposed a multi-task knowledge distillation method. By incorporating network-theory-based graph metrics as auxiliary tasks, we show on both synthetic and real datasets that the proposed multi-task learning method can improve the prediction performance of the original learning task, especially when the training data size is small. 

%In this work, we proposed a multi-task knowledge distillation method to improve graph-level representation in a semi-supervised fashion. By incorporating network-theory-based graph features as auxiliary tasks, we showed on both synthetic and real datasets that the proposed multi-task framework can improve the prediction performance of the original main task, especially when the training data size is small. We also showed that the proposed method outperforms or matches strong graph prediction baseline methods on two real benchmark datasets.

%Inspired by the work of knowledge distillation~\cite{hinton2015distilling}, we proposed a novel multi-task knowledge distillation framework for graph representation learning. By incorporating network-theory-based graph features as auxiliary tasks, the proposed framework helps the graph representation learning model learn better graph embeddings. We showed on synthetic datasets that the proposed multi-task framework can improve the prediction performance of the original main task comparing to the corresponding single-task graph representation learning model, especially when the training data size is small. We further showed on real benchmark datasets that the proposed method outperforms or matches state-of-the-art graph prediction methods.

\end{abstract}

\section{Introduction}
Graph structured data are ubiquitous in application domains of machine learning, such as social networks~\cite{szabo2010predicting, kupavskii2012prediction}, chemistry~\cite{wale2008comparison}, and biology~\cite{toivonen2003statistical, borgwardt2005protein}. However, learning from graph structured data is very challenging, as they cannot be directly represented in a tabular form, which is what most traditional machine learning models are designed for. Therefore, how to properly represent the graph structured data has been a key factor influencing the performance. 
%In this paper, we are particularly interested in graph-level prediction problems, which require a model to predict the label associated to the entire graph.

A common and straightforward method of graph representation is to calculate handcraft features from the graph based on domain knowledge to obtain a tabular representation. Though being successful in some applications~\cite{szabo2010predicting, kupavskii2012prediction}, this method has a heavy demand of domain expertise and suffers from information loss in the handcraft features. %Therefore it requires problem-specific designs for each new problem and might lack model capacity for complicated tasks. 
Graph representation learning methods~\cite{perozzi2014deepwalk, tang2015line, grover2016node2vec,kipf2016semi}, which learn a graph representation model automatically from data, overcome these defects but require a large amount of labels. While co-occurrence-based unsupervised learning methods are widely used in node-level or subgraph-level prediction problems (labels are associated with nodes or subgraphs), most existing graph-level (labels are associated with the entire graph) prediction models are trained with supervised learning. On the other hand, when successful, we know the machine learning models with handcraft graph features require much fewer labels to train, thanks to the domain knowledge provided by the features. And there is rich literature in network theory providing useful graph metrics. So it would be desirable if we can incorporate the existing domain knowledge from network theory into the graph representation learning to reduce the number of training labels required. 

% Key: combination of network theory and graph representation learning

\begin{figure*}[h]
	\vskip -15pt
	\centering
	\includegraphics[width=\textwidth]{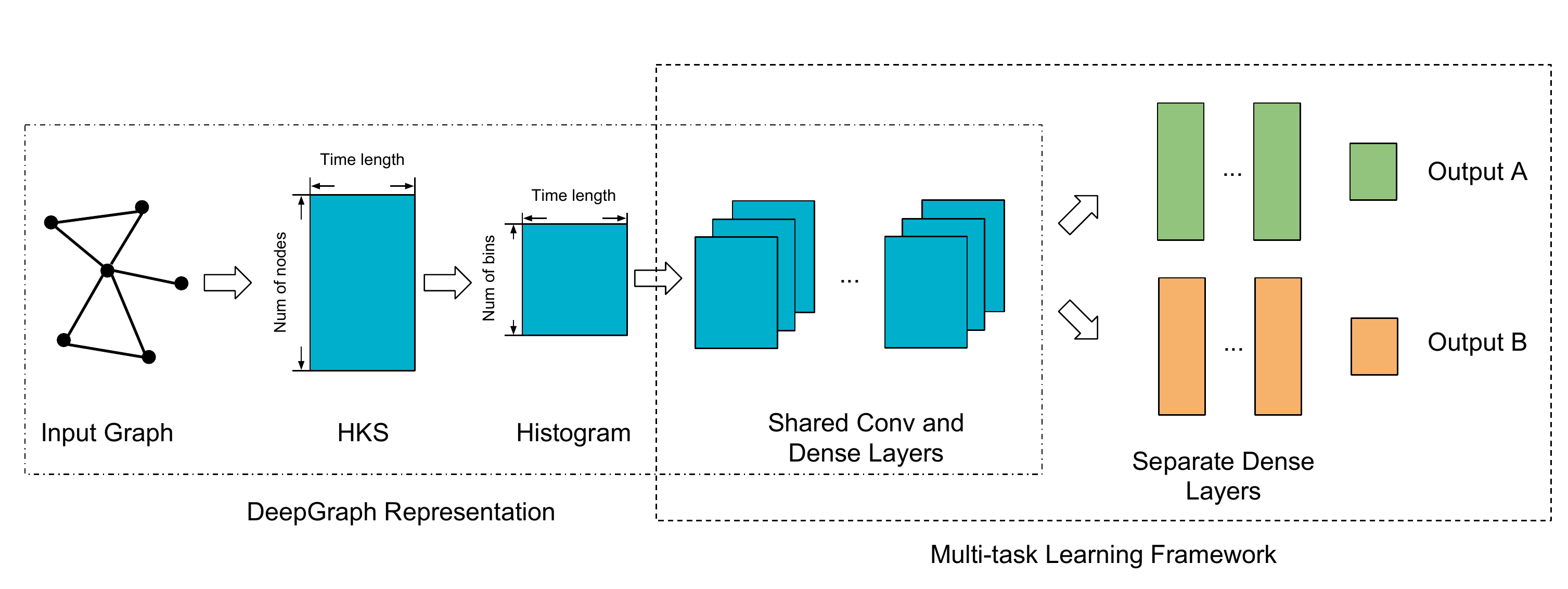}
	\caption{Implementation of the multi-task knowledge distillation method for graph representation learning.}
	\label{fig:model}
	\vskip -10pt
\end{figure*}

In this paper, we propose a novel multi-task knowledge distillation method for graph representation learning. We share an abstract view of knowledge with \citet{hinton2015distilling} that the knowledge can be represented as a mapping from input vectors to output vectors. Here we encode the knowledge of the network-theory-based graph metrics into the mapping from the raw graphs to the metric values. We then use these graph metrics as auxiliary tasks and distill the knowledge of network theory into the learned graph representations via multi-task learning. We implement the proposed method on top of DeepGraph~\cite{li2016deepgraph}, a recently proposed graph-level representation learning method, as a proof-of-concept. In general, however, the proposed method should be compatible with any graph representation learning models that are trained through supervised learning. We illustrate the implementation of the proposed method in Figure~\ref{fig:model}. 

Finally, we evaluate the proposed method on both synthetic datasets and real benchmark datasets. Experimental results show that the domain knowledge can improve the main task performance, especially when training labels are scarce.

\section{Approach}

In this section, we describe the proposed multi-task knowledge distillation method shown in details. The method contains two building blocks, a Graph Representation block, and a Multi-task Learning block.

\subsection{Graph Representation}

We use a similar graph representation structure as DeepGraph~\cite{li2016deepgraph} at the bottom of our model (shown in the left block in Figure~\ref{fig:model}). We first calculate the Heat Kernel Signature (HKS)\footnote{See Appendix~\ref{app:hks} for more details.} as a graph descriptor and then feed the histogram of the HKS into a convolutional neural network (CNN) to learn the representation of the graph. 

%In principle, we can plug in here any graph representation models that can be trained through supervised learning. We adopt DeepGraph, a high-order spectral-based graph representation model, as opposed to GCN~\cite{kipf2016semi} style low-order graph representation models, because high-order models may be able to capture more global graph patterns.

\begin{figure}[t]
\vskip -15pt
\centering
\includegraphics[width=0.49\textwidth]{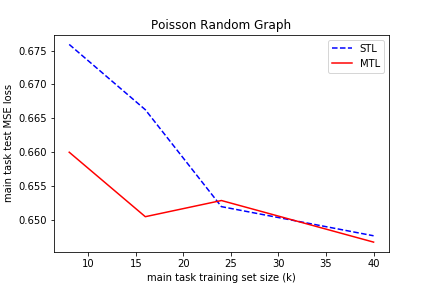}
\includegraphics[width=0.49\textwidth]{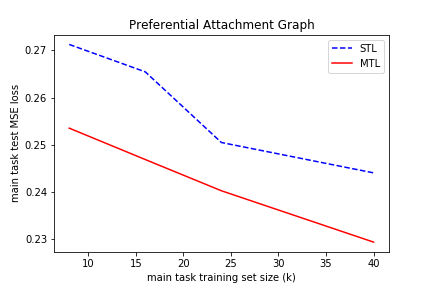}
\caption{Test MSE of the Diameter task vs different sizes of training data of the Diameter task on Poisson random graphs and preferential-attachment graphs.}
\label{fig:synthetic}
\vskip -15pt
\end{figure}

\subsection{Multi-task Learning}

Denote pairs of graph and (graph-level) label as $\{(G_i, y_i)\}_{i=1}^N$, where $G_i \in \mathcal{G}, y_i \in \mathcal{Y}$ and $\mathcal{G}, \mathcal{Y}$ are the spaces of all possible graphs and labels respectively. The graph-level prediction can be formulated as a supervised learning problem that finds a parameterized mapping function $f:\mathcal{G} \rightarrow \mathcal{Y}$ minimizing $J(\theta) = \frac{1}{N}\sum_{i=1}^N L(y_i, f(G_i;\theta)),$ 
where $L$ is a prediction error function and $\theta$ are the model parameters. 

In the proposed method, we use network-theory-based graph metrics as auxiliary tasks and train these tasks together with the main prediction task together through multi-task learning~\cite{caruana1998multitask}. In the multi-task setting, the label $y_i$ becomes a vector  $y_i = (y_1^1, y_i^2, \dots, y_i^K) \in \mathcal{Y} = \mathcal{Y}^1 \times \mathcal{Y}^2 \times \cdots \times \mathcal{Y}^K,$ 
and $K$ is the number of tasks. The problem becomes finding $K$ functions $f_k: \mathcal{G} \rightarrow \mathcal{Y}^k, k=1,2,\dots,K$ that minimizes 
$$\sum_{k=1}^K \frac{1}{|\mathcal{I}_k|} \sum_{i \in \mathcal{I}_k} \alpha_k L_k(y_i^k, f_k(G_i;\theta, \theta_k)),$$
where $\mathcal{I}_k \subset {1, 2, \dots, N}$ is the set of examples that have labels in task $k$ and $|\mathcal{I}_k|$ is the size of $\mathcal{I}_k$; $L_k$ is the prediction error function for task $k$; $\theta_k$ are the task-specific model parameters of $f_k$; $\theta$ are the model parameters shared by all $f_k, k=1,2,\dots,K$; $\alpha_k$ are hyper-parameters for task weights.  

We specify the model function for task $k$ as 
$$f_k(G;\theta, \theta_k) = g_k(h(G;\theta);\theta_k),$$
where $h(\cdot; \theta)$ is the DeepGraph representation and is shared by all the tasks and $g_k(\cdot; \theta_k)$ is the separate dense layers for task $k$. Typically, $g_k(\cdot; \theta_k)$ is a simple model (e.g., a linear model), so that the graph representation $h(\cdot; \theta)$ is forced to capture the knowledge from the auxiliary tasks.

\vpara{Network-theory based graph metrics as auxiliary tasks}

It is known that the auxiliary tasks should be related to the main task to improve the model performance~\cite{misra2016cross,ma2018modeling} but measuring task relatedness in general has been an open problem. For this specific scenario, however, the network-theory-based graph metrics are likely to be related because they have been useful as handcraft features in many graph-level prediction tasks~\cite{szabo2010predicting,kupavskii2012prediction}. 

Another advantage of using these graph metrics as auxiliary tasks is they are usually easier to obtain than the labels of the main task. We can calculate infinite number of auxiliary labels (e.g., from randomly generated graphs) so long as computation resources permit.

Finally, as the benefit of knowledge distillation, using the graph metrics as auxiliary labels rather than graph features helps reduce the computational cost at the test time, as we do not need to calculate these metrics for test data. This is especially important for applications in certain online services.

In this paper, we use Density and Diameter as our auxiliary tasks as a proof-of-concept. 

\section{Experiments}
In this section, we test the proposed method on both synthetic data and real data. 

\begin{figure}[ht]
	\centering
	\includegraphics[width=0.49\linewidth]{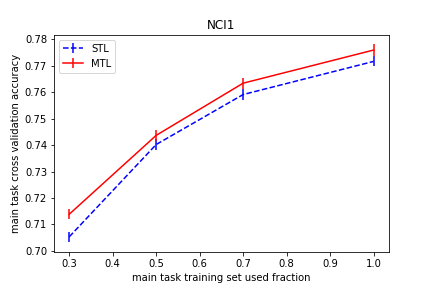}
	\includegraphics[width=0.49\linewidth]{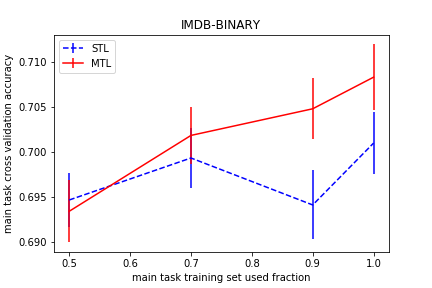}
	\caption{10-fold cross validation test accuracy of the main task vs the fraction of the training dataset used on NCI1 and IMDB-BINARY datasets. The error bar indicates the standard deviation of the mean of the cross validation experiments.}
	\label{fig:real}
	\vskip -10pt
\end{figure}

\subsection{Datasets}
\vpara{Synthetic data} We adopt two commonly used random graph models in network theory, Poisson random graph (a.k.a. Erd\H{o}s-R\'enyi model) and preferential attachment graph (a.k.a. Barab\'asi-Albert model), to generate our synthetic datasets. For each model, we generate 100k graphs and calculated both the Density and the Diameter as labels for each graph. The details of synthetic data generation can be found in Appendix~\ref{app:syn}.

\vpara{Real data} Two real graph-level prediction benchmark datasets, NCI1~\cite{wale2008comparison} and IMDB-BINARY~\cite{yanardag2015deep}, are used. More details can be found in Appendix~\ref{app:real}.

\subsection{Experiment Setup}
For each dataset, we first split the whole dataset into training, validation, and test dataset with the proportion 8:1:1. We vary the number of labels available for the main task. For synthetic datasets, we set the main task as Diameter and the auxiliary task as Density because Diameter has a higher computation complexity. For the real datasets, we set the original prediction task associated with the dataset as the main task, and both Diameter and Density as the auxiliary tasks. 

We implement both the multi-task model and the single-task model as a neural network of two convolutional layers followed by two fully connected layers. For multi-task model, the two convolutional layers and the first fully connected layer are shared by two tasks while the second fully connected layer is separated. For each setting, we tune the hyper-parameters for the multi-task model and the single-task model separately using random search on the validation set (see Appendix~\ref{app:hp}). We evaluate the models on the independent test set. 

For the real datasets, because the data sizes are small, we use 10-fold cross validation to evaluate the models. We only tune the hyper-parameters on the first fold and fix them for other folds.

\subsection{Results}

\vpara{Synthetic data}
The results of Poison random graph and preferential attachment graph are shown in Figure~\ref{fig:synthetic}. As can be seen in the results, for both datasets, the single-task model gets lower MSE on the test set when the training data size increases. With the help of the auxiliary task (Density task), the multi-task model is almost always better than the single-task model on the synthetic datasets. Yet the gap between the performance of the multi-task model and that of the single-task model is larger when the training data size is relatively small. This trend is especially obvious on the Poison random graph dataset, where the gap vanishes as the training data size increases to 24k. 

%It is also interesting that the MSE score goes up a little bit when the training data size goes from 16k to 24k, which indicates the performance of the multi-task model has a more complex pattern than the single task model. 

\vpara{Real data}
The results of the multi-task model and the single-task model with variable training data size on NCI1 and IMDB-BINARY are shown in Figure~\ref{fig:real}. The result of the NCI1 dataset is consistent with what we observed in the synthetic data: the multi-task model is better than the single-task model, and the gap between the multi-task model and the single-task model is larger when the training data size is smaller. While the multi-task model is also mostly better than the single-task model on the IMDB-BINARY dataset, the gap between the multi-task model and single-task model does not show the same pattern as in other datasets. This is likely because the IMDB-BINARY dataset is too small therefore there is high variance in the results. Indeed, the error bars of the the evaluation scores indicate that the difference between the two models, especially when the fraction of the training dataset used is small, is not significant. %It is also abnormal that the accuracy of the single-task model drops when the fraction of training dataset used goes from 0.7 to 0.9. This is potentially also because the high variance of the evaluation results. 

\section{Conclusion}
In this paper, we propose a novel multi-task knowledge distillation method for graph representation learning. This method incorporates network-theory-based graph metrics as auxiliary tasks via multi-task learning to learn better graph representations, and we have demonstrated that the proposed method improves the quality of the main prediction task. In future work, we plan to adopt more advanced graph representation models to further optimize for the performance on real datasets.

%We tested our hypothesis that the fewer labels available in the main task training data, the larger the multi-task model will gain comparing to the corresponding single-task model on the synthetic datasets. We also showed the proposed method outperforms or matches the state-of-the-art graph prediction methods on real benchmark datasets. 

% \newpage
\bibliography{bibliography}
\bibliographystyle{apalike}

\newpage
\appendix
\section{Experiment Details}

In the experiments, we want to test our hypothesis that the multi-task model can help the main task when the number of labels available for the main task is small, and we found some difficulties in using the real datasets for this hypothesis testing. 

First, We would like to conduct control experiments varying the size of training dataset of the main task. However, for each individual dataset and a given data size $n$, it is difficult to determine which ``phase'' the dataset resides in: $n=100k$ might still be too small for a difficult task while $n=1000$ could be large enough for a simple task; and we do not know difficulty of the main task associated with the dataset as a priori. Moreover, the real benchmark graph prediction datasets we got are in the level of thousands of examples. The test set is so small that the evaluation score on the test set has a high variance. As a consequence, we have to run heavy cross validation experiments before we get results that are significant. 

Therefore to overcome these difficulties, we used synthetic data generated by random graph models as a proof-of-concept testing of our hypothesis. In synthetic data, we can easily control the data size and have abundant test data to have robust evaluation.

\subsection{Datasets}
\subsubsection{Synthetic Data}
\label{app:syn}
\begin{itemize}
	\item Poisson random graph: For each graph, the number of nodes $n$ is randomly generated from a normal distribution $\mathcal{N}(30, 10^2)$ with the minimum $5$. The link probability $p$ is randomly generated from a normal distribution $\mathcal{N}(0.3, 0.12^2)$ with the minimum $0.05$. 
	\item Preferential attachment graph: For each graph, the number of nodes $n$ is randomly generated from a normal distribution $\mathcal{N}(30, 10^2)$ with the minimum $5$. The number of edges $k$ for each new node is randomly generated from a normal distribution $\mathcal{N}(6, 2^2)$ with the minimum $2$ and maximum $n-3$.
\end{itemize}

\subsubsection{Real Data}
\label{app:real}
\begin{itemize}
	\item \textbf{NCI1}~\cite{wale2008comparison} are chemical compounds screened for activity against non-small cell lung cancer and ovarian cancer cell lines. There are 4110 graphs in this dataset.
	\item \textbf{IMDB-BINARY}~\cite{yanardag2015deep} is a movie collaboration dataset. The nodes of each graph represent actors and two nodes are linked by an edge if the corresponding actors appear in the same movie. Each graph is the ego-network of an actor, and the task is to predict which genre an ego-network belongs to. There are 1000 graphs in this dataset. 
\end{itemize}

\subsection{Hyper-parameter Tuning}
\label{app:hp}

For both synthetic data and real data, the hyper-parameters we tuned are listed below.

\begin{itemize}
	\item Number of time step for HKS:  Uniformly from \{16, 32, 64, 128\}.
	\item Minimum time step for HKS: Uniformly from $[e^{-6}, e^{1}]$ in log scale.
	\item Maximum time step for HKS: Uniformly from $[e^2, e^6]$ in log scale.
	\item Number of bins for HKS histogram: Uniformly from \{16, 32, 64, 128\}.
	\item Kernel sizes of the convolutional layers: Uniformly from \{3, 5, 7\}.
\end{itemize}

The first fully connected layer size is set as 60 and the second fully connected layer size is set as 40. Adam \cite{kingma2014adam} algorithm with the default hyper-parameter setting is used to train both models and early stop is used with the validation set. 

For real data, there is one more hyper-parameter: which auxiliary task to use (either diameter or density). 

\subsection{Comparison with Strong Baselines}

We also compare the proposed methods with Strong baseline methods on the full dataset.

\vpara{Comparison methods}

\begin{itemize}
	\item \textbf{DGMTL} 
	\item \textbf{Deep graph kernels}~\cite{yanardag2015deep} are a group of deep variants of some graph kernels, which are reported to outperform or match the corresponding graph kernels on these benchmark datasets. In this paper, we compare with the deep variants of the graphlet kernel (\textbf{Deep GK}) and the Weisfeiler-Lehman subtree kernel (\textbf{Deep WL}). 
	\item \textbf{PSCN}~\cite{niepert2016learning} is an extension of CNN to graphs, which first determines an ordered sequence of nodes from the graph, then generates a neighborhood graph of constant size for each node in the sequence, and generates a vector representation (patch) for each neighborhood so that nodes with similar structural roles in the neighborhood graph are positioned similarly in the vector space, finally feeds the patches to a 1D CNN.
\end{itemize} 

As we use the same setting as of~\citet{yanardag2015deep, niepert2016learning}, we just report the results from those papers for comparison. ``N/A'' means the result is not available from the original paper.

\vpara{Results} The results are shown in Table~\ref{tbl:real}. For the NCI1 dataset, the Deep WL method still remains the best, probably because the WL kernel is a very good fit for this specific task. Our approach (DGMTL) outperforms all other methods. For the IMDB-BINARY dataset, while the PSCN method has a slightly higher mean accuracy than our approach (DGMTL), it has very high variance. Overall, the proposed method is comparable with the strong baseline methods. 

%\begin{table*}[t]
%\caption{10-fold cross validation test accuracy of state-of-the-art graph prediction methods and our approach on the real benchmark datasets.}
%	\begin{tabular}{|c|c|c|c|c|c|c|}
%		\hline
%		  & Deep GK & Deep SP & Deep WL & PSCN & DGSTL & DGMTL\\
%		\hline
%		NCI1 & 62.48 ($\pm$ 0.25) & 73.55 ($\pm$ 0.51) & 80.31 ($\pm$ 0.46) & 76.34 ($\pm$ 1.68) & 77.17 ($\pm$ 0.19) & 77.56 ($\pm$ 0.21)\\
%		\hline
%		IMDB-B & 66.96 ($\pm$ 0.56) & N/A & N/A & 71.00 ($\pm$ 2.29) & 70.10 ($\pm$ 0.35) & 70.83 ($\pm$ 0.36)\\
%		\hline
%	\end{tabular}
%	\label{tbl:real}
%\end{table*}

\begin{table*}[t]
\caption{10-fold cross validation test accuracy of state-of-the-art graph prediction methods and our approach on the real benchmark datasets.}
	\begin{tabular}{|c|c|c|c|c|c|}
		\hline
		  & Deep GK & Deep WL & PSCN & DGSTL & DGMTL\\
		\hline
		NCI1 & 62.48 ($\pm$ 0.25) & 80.31 ($\pm$ 0.46) & 76.34 ($\pm$ 1.68) & 77.17 ($\pm$ 0.19) & 77.56 ($\pm$ 0.21)\\
		\hline
		IMDB-B & 66.96 ($\pm$ 0.56) & N/A & 71.00 ($\pm$ 2.29) & 70.10 ($\pm$ 0.35) & 70.83 ($\pm$ 0.36)\\
		\hline
	\end{tabular}
	\label{tbl:real}
\end{table*}

\section{Heat Kernel Signature (HKS)}
\label{app:hks}
HKS maps isomorphic graphs to the same representation and graphs with the same HKS representation are isomorphic. Therefore HKS is a very informative graph descriptor.

Given a graph $G=(V, E, W)$, where $V, E$ are the set of nodes and edges of $G$ respectively and $W \in \reals^{|V|\times |V|}$ is a matrix denotes the edge weights between every pair of nodes. The heat kernel $h_z(i, j)$, a function of two nodes $i, j \in V$ at any given diffusion step $z$, represents the amount of aggregated heat flow through all edges among the two nodes after diffusion step $z$. The heat kernels are computed using eigenfunction expansion of a graph Laplacian. The graph Laplacian is defined as: 
$$L = D - W,$$ 
where $D$ is a diagonal degree matrix with diagonal entries being the summation of rows of $W$: $D_{ii} = \sum_j w_{ij}$.

The heat kernel is then defined as
$$h_z(i, j) = \sum_{k=1}^{|V|} e^{-\lambda_kz}\phi_k(i)\phi_k(j),$$
where $\lambda_k$ is the $k$-th eigenvalue of $L$ and $phi_k$ is the corresponding eigenfunction. 

In practice, using heat kernels as representation of graphs has a high computation complexity so heat kernel signature was introduced as a computationally cheaper representation of graphs. Specifically, the heat kernel signature is a matrix $H \in \reals^{|V| \times T}$ such that
$$H_{ij} = h_{z_j}(i, i),$$
where $z_j \in \{z_1, z_2, ..., z_T\}$ is a time step from a finite set of $T$ diffusion steps. 

Finally, the heat kernel signature is summarized into a histogram $S \in \reals^{B \times T}$, where $B$ is a hyper-parameter representing the number of histogram bins, and then is fed into a CNN.

For more details of the graph representation, we refer the readers to the paper~\cite{li2016deepgraph} for the DeepGraph model and the paper~\cite{sun2009concise} for the original HKS literature. 

\end{document}